\documentclass[journal]{IEEEtran}
\usepackage{cite}

\ifCLASSINFOpdf
  \usepackage[pdftex]{graphicx}
\else
  \usepackage[dvips]{graphicx}
\fi
\usepackage{amsmath}
\newtheorem{definition}{Definition}
\newtheorem{theorem}{Theorem}
\usepackage{array}
\usepackage{caption}

\begin{document}
%
\title{The Unreasonable Effectiveness of Deep Learning}
%
%
%

\author{Dr. Finn Macleod, Beautiful Data
\thanks{Manuscript received March 18, 2018}}

%
%

\markboth{IEEE Transactions on Neural Networks and Learning Systems}%
{Shell \MakeLowercase{\textit{et al.}}: Bare Demo of IEEEtran.cls for IEEE Journals}
%



\maketitle

\begin{abstract}
We show how well known rules of back propagation arise from a weighted combination of finite automata. By redefining a finite automata as a predictor we combine the set of all $k$-state finite automata using a weighted majority algorithm. This aggregated prediction algorithm can be simplified using symmetry, and we prove the equivalence of an algorithm that does this. We demonstrate that this algorithm is equivalent to a form of a back propagation acting in a completely connected $k$-node neural network. Thus the use of the weighted majority algorithm allows a bound on the general performance of deep learning approaches to prediction via known results from online statistics. The presented framework opens more detailed questions about network topology; it is a bridge to the well studied techniques of semigroup theory and applying these techniques to answer what specific network topologies are capable of predicting. This informs both the design of artificial networks and the exploration of neuroscience models. 
\end{abstract}

%
\IEEEpeerreviewmaketitle

\section{Introduction}
%
%
%
%
\IEEEPARstart{I}{n} applied prediction and data science challenges with large volumes of data and little in the way of \lq feature engineering\rq  by an expert human, techniques of deep learning have dominated other approaches. However, it is not yet possible to define generally the best network topology for a given data set and bound the error rate. 

In an applied prediction setting, the "best" prediction model is often undiscovered because the space of potential models outpaces the computational resources to search through it. One almost never has the computational power to search the full model space. In this setting, we're interested in how to build an \textit{efficient} prediction system; a system that predicts optimally, with a restricted amount of resources. 

Consider predicting a sequence $a = a_1 a_2 \ldots $ where every element is drawn from a finite set $A$ with $|A|$ elements,  $\{a^1 \ldots a^{|A|}\}$. The prediction functions, $\alpha$, are every possible function mapping a past set of $t$ elements into a possible future : 
\begin{equation}
a_1 a_2 \ldots a_t \xrightarrow {\alpha}  a_{t+1} 
\end{equation}
So for a sequence of length $t$, there are 
\begin{equation}
    |A|^{|A|^t}
\end{equation} 
such functions. The number of functions (or models) mapping historical data to the future grows doubly-exponentially as historical data grows with time. This double-exponential growth quickly outpaces computational resources to search through all possible predictors, and compels us to be more selective in our search for the best model. 

To formalise this setting of restricted resources, we define an index on predictor functions with a finite number of internal states; we refer to the the number of internal states as the \textit{complexity} of a given predictor.

We will show accuracy bounds for sets of predictors with a given complexity via well known results of consensual learning theory and online statistics. In particular we will use the weighted majority algorithm, known in its more general form as the aggregating algorithm. The aggregating algorithm combines the predictions of a set of functions in a provably optimal way. These are not absolute bounds, as in classical probability which would require assumptions of stationarity and ergodicity, but rather the bounds of online statistics - that our prediction accuracy will be within a certain distance of the best predictor in a set. 

More specifically, we define prediction error using a loss function $L(A,A) \rightarrow  \{0,1\}$, for example
\[
L(a_{t+1}, \alpha(1 \ldots a_t)) = 
\begin{cases}
    0,& \text{if identical}\\
    1,              & \text{otherwise}
\end{cases}
\]
but there are many alternatives. We are interested in the cumulative loss over time 
\begin{equation}
E(\alpha) = \sum_{i=1}^t L(a_{i+1},\alpha(1 \ldots a_i)).
\end{equation}
The weighted majority algorithm and its generalisation, the aggregating algorithm define a prediction strategy with bounds on this cumulative loss for a wide class of loss functions. When combining $N$ predictors, these bounds are
\begin{equation}
E(Agg) < c_1(L)E(\alpha) + c_2(L)\log N
\end{equation}
for all $\alpha$ in the pool of N predictors. $c_1$ and $c_2$ are constants for a given loss function  - a variety of which are calculated in the paper by vovk \cite{Vovk2}, along with a more thorough introduction to the approach of online statistics.

The above equation shows that the prediction accuracy of the aggregated predictors will be within a specified bound of the performance of the best predictor, over any sequence. More importantly, In some ways, the aggregating algorithm can be thought of as a generalisation of the bayesian mixture, and often performs much better than these worst case bounds. See \cite{Vovk2} for a full discussion. 

We present a simple definition of the aggregating algorithm for the above loss function:\begin{definition}[Weighted Majority Algorithm]
 Each prediction function, $\alpha$, is assigned an initial weight, $w_{\alpha} = 1$. At time $t$ once calculates for each $a_i \in A$
 \begin{equation}
 \sum_{\alpha(a_1 \ldots a_t) = a_i} w_{\alpha}
 \end{equation}
 The prediction of the aggregating algorithm is the element of $a_i$ which corresponds to largest of these quantities. Then for each $\alpha$ that predicted correctly, the weight $w_\alpha$, is multiplied by a constant factor $\lambda >1$, and the process repeated. In practical applications there is a normalisation step to avoid the weights becoming overly small. 
 \end{definition}

Here, we apply the aggregating algorithm to the set of all predictors with complexity $K$. This immediately gives us weak bounds for arbitrary input. This algorithm will perform within aggregating bounds of the best $K$-state finite state predictor over any sequence, although the number of predictors combined, $N$, is very large.

The calculation of this algorithm is a significant undertaking, even for small $K$. However, there are a vast amount of symmetries present as we are calculating many similar operations over the complete set of \textit{all} predictors of a given complexity. This allows us to to simplify the algorithm vastly. We show a simplified, yet equivalent method of calculating the aggregating algorithm applied to all predictors of complexity $K$, has the same rules as the familiar back propagation algorithm on a completely connected neural network with $K$ nodes.

Consider the analogy: finite predictors are individual atoms, and a neural network represents the macroquantities of pressure, volume and temperature; we present the mechanics of how they are connected.

\section{Prediction using finite state automata}

\begin{figure}[!t]
\centering
\includegraphics[width=2.5in]{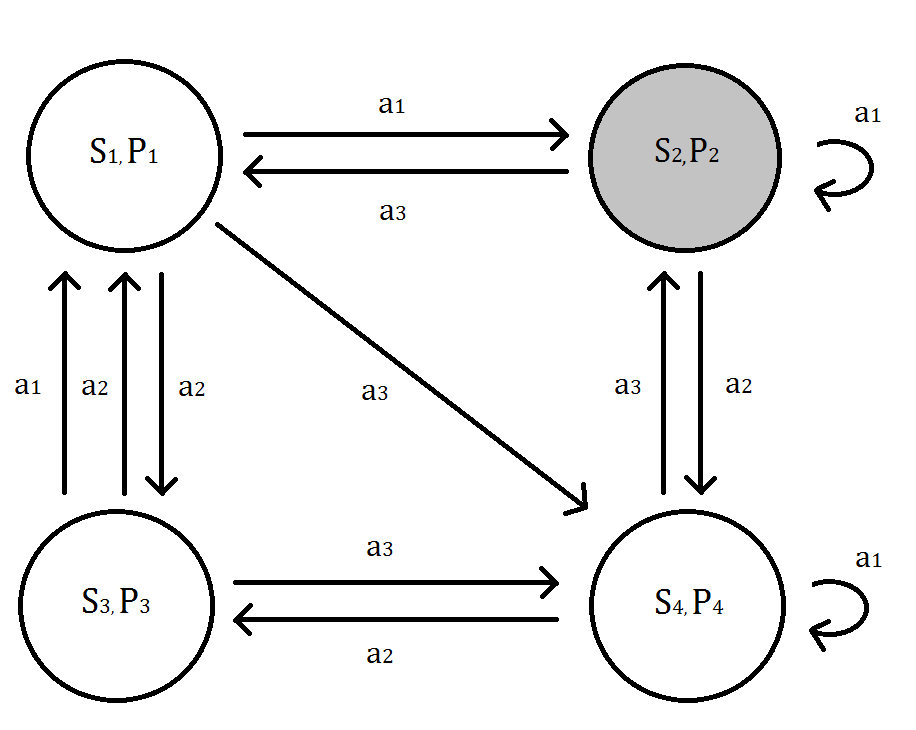}

\caption[content]{Example of a four state predictor, with input alphabet, $|A|$ of size 3, and matrix
\newline
$
G = \left( \begin{matrix}
    s_2 & s_2 & s_1 & s_4\\
    s_3 & s_4 & s_1 & s_3\\
    s_4 & s_1 & s_4 & s_2\\
\end{matrix}\right)
$
\newline
The active state is shaded in grey. We combine the set of all configurations of such finite automata style predictors for a given number of states}
\label{fig_sim}
\end{figure}

We redefine the traditional concept of a finite automata, into that of a predictor, by defining the output of each state as the prediction for the next input:

\begin{definition}[The finite state Predictor]
A $K$ state predictor, with input alphabet $A$, consists of 
\begin{itemize}
\item a set of $K$ internal states, labelled $s_1, \ldots, s_K$
\item a prediction associated with each state, $p_i \in A$
\item a designated state, $s^{active}_t \in S$, known as the active state at time $t$
\item an $|A|\times K$ matrix, G, with entries, $G_{ij} \in S$. Together with an element from the input sequence the matrix $G$ defines the next active state in time; If $s^{active}_t = s_j$ and with input $a_i$, then
\begin{gather}
s^{active}_{t+1} = G_{ij}.
\end{gather}
\end{itemize}
\end{definition}

The set of all predictors with complexity $K$ represents all possible configurations of finite state predictors with $K$ internal states. We note that the structure presented is identical to that of a finite state automata, defined to function as a predictor. Thus the well studied semigroup theory associated with finite state automata (for example, see \cite{TMST69}) has an immediate connection to that of prediction using back propagation.

\section{Relative Complexity}

The finite state predictors allow us to frame an introductory question of relative complexity:
\begin{quotation}
\textbf{"How accurately can a finite state predictor with $K$ states predict a periodic sequence
of period $n$?"} 
\end{quotation}
When $K \geq n$ there is a $K$-state predictor that can predict the sequence perfectly, however, when $n > K$, there is necessarily error. We're looking for the shape of the graph in Fig 2,
\begin{figure}[!t]
\centering
\includegraphics[width=2.5in]{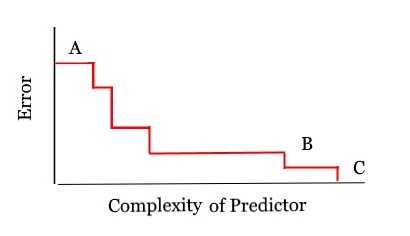}
\caption{Periodic Input of length $n$: The best a 1-state finite state predictor can do is guess the most frequent state of the input sequence (A). With $n-1$ states (B) there is at most one error. With $n$ states, it is possible to achieve zero error (C)}
\label{fig_profile}
\end{figure}
which requires us to have a method of calculating the best $K$-state predictor for any $n$. One method of solution is an algorithm that will track all $K$-state predictors, define an index which records the errors for each predictor, and thus we can pick the best one. There are
\begin{eqnarray}
K^{|A|K}|A|^K K,
\end{eqnarray}
finite state predictors -  $K^{|A|K}$ matrices, each with $|A|^{K}$ different variations in prediction, and $K$ possibilities for active initial states. However this set of all finite state predictors contains a lot of symmetry. For example, there are a number of predictors with identical matrices, which make opposite predictions at each state. These opposite pairs of predictor can be tracked simultaneously which reduces the number of calculations.

This problem is nearly identical to the combination of $K$-state predictors using the aggregating algorithm, with the index mentioned above being analagous to the weight of a predictor. Indeed solving the problem for the weighted majority algorithm defines an upper bound on this problem.

In addition to the finite period case, combining predictors using the aggregating algorithm defines bounds of accuracy for arbitrary input sequences. In this paper, we show how using the many symmetries present in the set of all $K$ state automata can vastly simplify this task of simultaneously calculation.

\setlength{\parindent}{5ex}

\section{Revealing Backpropagation}

Backpropagation rules in neural networks come in a number of flavours. Here we take a very simple linear case. 

\begin{definition}[simple backpropagation]
We consider a completely connected network, consisting of $K$ nodes, each node has an active weight, $w_i$, and connection weights between the nodes, $w_{ij}$. At each timestep, the active weight travels from node $i$ to node $j$ with weight proportional to the connection weights, specifically $w_i w_{ij}$. Each input element of the alphabet of the input sequence is associated with a node, when that element is recieved, the active weight at that node is updated by a factor $\lambda$. In addition, the connection weights travelling to this activated node are updated by a factor $\lambda$. Predictions are proportional to the active weight at each node before any updates. The node with the most weight is the most likely, and is the output prediction of the network.
\end{definition}

\begin{theorem}
We show that the predictions made by the aggregating algorithm applied to the set of all finite state predictors with $K$ states can be calculated with a set of $K^K + K$ quantities, provided $|A|\leq K$. The structure of this algorithm behaves with the rules of simple backpropagation.
\end{theorem}

We note the following symmetries
\begin{enumerate}
\item Transition symmetry: For every group with a transition $ij$ under input state $s_t$ say, there are $K-1$ other finite state predictors that are identical apart from transition $ij$ replaced by transition $ik$ for all $k \neq j$. True for each input state.
\item Prediction symmetry. For every group that predicts $p_i$ at $s_j$ there is a group with identical $G$ that predicts $p_k$ at $s_j$ for all $i,j,k$.
\end{enumerate}

\subsubsection*{Prediction Classes}

For a given finite state predictor, we refer to the set of $p_i$'s as a vector, $\vec{p}$. Then

\begin{definition} A prediction class, $p(G)$, consists of all finite state predictors that share the same transition matrix and active states - identical $G$ and $s^{active}$, but all possible variations of $\vec{p}$
\end{definition}
We note that there are $|A|^K$ finite state predictors in a prediction class. We can write the total sum of their weight as:
\begin{equation}
    \sum_{i_1 = 1}^{|A|} \ldots \sum_{i_K = 1}^{|A|} w(\vec{p})
\end{equation}
One can represent this as a hypercube of predictors, with $K$ axes, each corresponding to a state, and the co-ordinates of a predictor are an index of the predictions it makes.

The action of updating weights for a prediction is represented on this hypercube by multiplying the weight at each co-ordinate in a given plane by the update factor $\lambda$ and then normalising all elements of the cube. When performing this update action on given plane in the hypercube, the total weight of planes along any other dimension are constant. 
\begin{figure}[!t]
\centering
\includegraphics[width=2.5in]{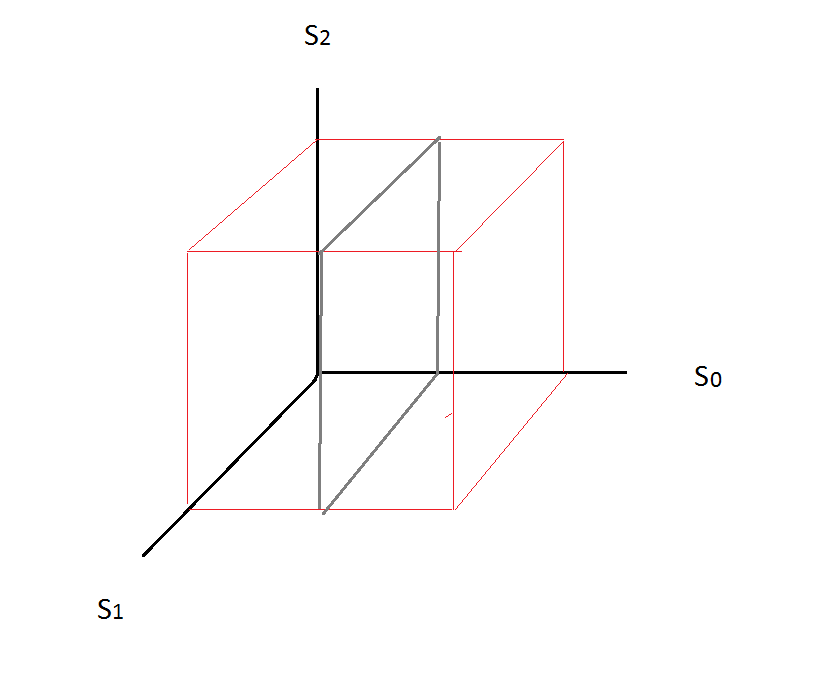}
\caption{Updating all predictors that predict $s_1$ at state $s_0$, is the action of multiplying all predictors in the plane by a factor $\lambda$ and normalising the cube. Planes along other dimensions of the cube will have constant weight under this process. We can calculate the aggregating algorithm by keeping track only of the total weight of each plane, thus reducing the number of quantities required from $|A|^K$ to $|A|K$}
\label{fig_hypercube}
\end{figure}

\setlength{\parindent}{5ex}
It is possible to define a reduced set of independent quantities to represent the quantities necessary to calculate the prediction of the aggregating algorithm when applied to a given prediction class. We call these quantities the transition weights: 
\begin{definition}
We define the transition weights of a prediction class:
\begin{equation}
w_{ij}(G) := \sum_{\vec{p}, p_i=s_j} w(\vec{p})
\end{equation}
\end{definition}
If $a_j$ is the next item in the input sequence after active state $s_i$ then system evolution is performed by: $w_{ij} \rightarrow \lambda w_{ij}$, and other prediction weights are untouched.

The aggregating algorithm applied to all finite state predictors can be calculated by keeping track of the absolute weight of each prediction class as well as the normalised $w_{ij}$ values. This is done by a vector $\vec{w(G)}$, where each $w_i(G)$ corresponds to the weight active at that state - for a single transition class, this vector is zero apart from one entry which is $w^{active}$, the total weight of the transition class. It evolves as:
\begin{equation}
G|a_i \hspace{1mm} \vec{w}_t(G) = \vec{w}_{t+1}(G)
\end{equation}
Where $G|a_i$ is the transition matrix formed by G under input $a_i$. Updates to the weight are determined by the ratio of weight predicting correctly. For a single transition class, $w(G)=w^{active}$. If $a_i$ is the input, then
\begin{eqnarray}
w^{active} \rightarrow & \lambda w_{ij}w^{active} + (1-w_{ij})w^{active}\\
=& w^{active} + (\lambda -1)w_{ij}w^{active}
\end{eqnarray}
In addition at each timestep we update the $w_{ij}$ by $\lambda$ if state $i$ is active according to the input.

\subsubsection*{Induction}
We can simplify the calculation of the aggregating algorithm further using transition symmetry to remove direct dependency on the input $a_i$ and average out the structure of each individual $G$ from the calculation. More specifically,
\begin{definition}
The transition weights of a $K$ state network are:
\begin{equation}
    w_{ij} = \frac{1}{N}\sum_G w_{ij}(G)
\end{equation}
The active weights of a $K$ state network are:
\begin{equation}
    w_i = \sum_{G} w_i(G).
\end{equation}
\end{definition}
These quantities are sufficient to calculate the aggregating algorithm applied to all K-state finite state predictors. We hypothesise that these quantities are sufficient to determine the calculation at the next timestep using the following evolution rules:
\begin{definition}[K-state system Evolution Rules]
First under input $a_j$
\begin{equation}
    w_j^{t} \rightarrow \lambda w_j^t
\end{equation}
Then
\begin{equation}
    w^{t+1}_j = \sum_i w^t_{ij}w^t_i
\end{equation}
under input $a_j$. Finally, we update the transition weights:
\begin{equation}
    w^{t+1}_{ij} = \lambda w^t_{ij},
\end{equation}
under input $a_j$, for all $i$. Otherwise constant in time.
\end{definition}

We prove true using induction. The hypothesis is true by transition symmetry when all weights are equal. We now suppose true for a given set of $w_i, w_{ij}$, and prove by induction for all updates.

\begin{figure}[!t]
\centering
\includegraphics[width=2.5in]{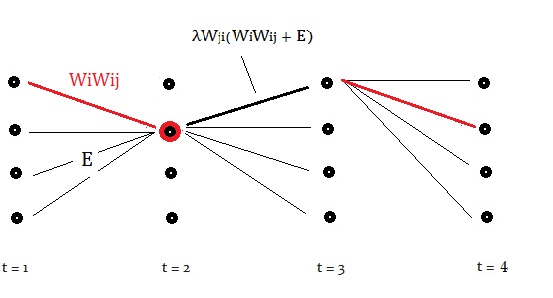}
\caption{Total weight transitioning to node $j$ can be considered in two parts: $w_i w_{ij}$ plus the rest, $E$. After an update both parts are distributed to all nodes according to the values of $w_{jk}$. Weight returning to node $i$ behaves in two ways. $\lambda w_{ji}E$ is equally distributed according to $w_{ik}$ values. The second part, $\lambda w_{ji}w_iw_{ij}$, entirely transitions to $j$. Thus an update can be modelled by updating $w_j$ by $\lambda$ and the ratio, $w_{ij}$ also by $\lambda$.}
\label{fig_returningweight}
\end{figure}

\setlength{\parindent}{5ex}
Total weight moving from $s_i$ to $s_j$ is $w_iw_{ij}$. Suppose input $a_j$. Then all weight at $s_j$ is updated by $\lambda$, confirming the first evolution rule.
\begin{equation}
w_j = \sum_{i=1}^K \lambda w_i w_{ij}
\end{equation}
This weight is then distributed amongst the new states according to the ratio given by the $w_{jk}$ for all $k$. Thus the amount returning to $s_i$ is 
\begin{eqnarray*}
w_{ji}w_j &=& w_{ji}\sum_{l=1}^K \lambda w_l w_{lj}\\
&=& w_{ji}\lambda w_iw_{ij}+  w_{ji}\sum_{l=1, l\neq i}^K \lambda w_l w_{lj}
\end{eqnarray*}
Thus, in addition to updating the active weight by $\lambda$, we update the $w_{ij}$ by $\lambda$. This can be seen by inspecting the two parts of the above equation which correspond to an update to the ratio $w_{ij}$ and the active weight. Most weight is distributed according to the ratio $w_{ij}$. However, the weight originating from $s_i$ will all transition to $j$ and only $j$ on the next step.Thus we update the ratio by $\lambda$

By updating in this way, we show that we have a greatly reduced set of quantities that can keep track of the system, and that these quantities obey the rules of simple back propagation.

We believe there may be several generalisations; to the non-linear case; to work off a gradient difference (rather than having a symmetric exchange of both a $w_{ij}$ and a $w_{ji}$); and the case of the non-deterministic aggregating algorithm.

\section{Discussion}

We believe that a precise understanding of what nodes and connections represent, will enable more precise training (eg. a specific part) and design of networks.

Neural nets are connected in many ways other than a completely connected graph. These connections reveal their ability to predict certain data sources. Evaluating these structures and comparing them to their performance on specific types of sequence can be approached via the well studied area of semigroup theory; certain "prime" groups will be able to predict certain sequences with a similar structure.

An intuitive approach to this is to design input sequences by a similar methodology; we imagine an unbiased random input stream running through a specific finite state automata to create a data sequence with a complexity upper bound. If one considers the classification of finite groups, one can pick out something somewhat esoteric eg. a  sequence generated by the monster group. This sequence would be impossible to predict with a high degree of accuracy unless the network topology met certain criteria of size and connectedness.

This framework also has application to the structure of the biological networks. We assume that evolution has optimised these networks to do two things well, both conserve energy and maximise prediction accuracy. Prediction accuracy is a slightly different quantity from memory; rememembering the past is different from being able to predict the future. The differences between exact memory recall and accurate prediction are subtle; any predictive model necessarily captures past data (remembers, if you will) but, under restricted resources, it is sometimes better to have an \lq inaccurate\rq model of the past. An analogy: One stores the "line of best fit" and not the "data points". 

This bending of "memory" to fit new information, is exactly the kind of behaviour one might expect in a well designed predictive system. Thus with regard to human memory, we like to view this property as not some kind of weakness, but a necessary quality of a system optimised to predict the future efficiently.

A finite state automata (and thus also aggregated sets of predictors) can store patterns that are infinitely far back in the past without needing to have an infinite memory (see Appendix 1). Forgetting and remembering a given piece of information is then directly linked to the predictive power of that information. The fading of memory with time is not a fallibility, it is the hallmark of a system optimised to predict the future whilst minimising resources. 

Professor Alexei Pokrovskii told me about unpublished experiments performed in Moscow where students were asked to generate random sequences and another group of students asked to build prediction algorithms. Whilst it is well known that humans have difficulty generating truly random sequences, the best predictors of these sequences were found to be finite state automata of 6 or 7 states. Experimentally, one could compare the connectivity of the regions involved with the number of groups and structure. Certain patterns would be unable to be generated by nets of a certain degree of connectedness, and vice versa, certain sequences would be unable to be predicted past a certain degree of accuracy by nets of a given degree of connectedness.

\appendices

\section{Recalling Significant Events Infinitely Far Back}
\newtheorem{example}{Example}
\begin{example}
Consider a finite state predictor comprised of two smaller finite state predictors A and B. The connection between them has a specific property: the only way of the active state passing from A to B is by a specific input sequence, $s_1 s_1 s_1 s_1 s_1$, say. Similarly, the only way of the active state passing from B to A is by the same input sequence. Thus if the active state is in section A it is guaranteed to remain in A until that specific sequence appears as input. Thus the predictions made by the finite state predictor depend on the starting state and the number of times the sequence $s_1 s_1 s_1 s_1 s_1$, has occurred. Thus the predictions are dependent on a property of the sequence which is not dependent on a finite window of time - one may have to look arbitrarily far back in the past to find the last instance of $s_1 s_1 s_1 s_1 s_1$,. The prediction of a Decision Tree predictor however, can always be determined by examining a fixed finite number of input digits back in the past.
\end{example}

More generally, one should choose to build ones prediction model from an ordered list of the most \textit{significant} state transitions in the input sequence, and not limit the use of ones finite memory to a time window.
\section*{Acknowledgment}

The author would like to thank Vladimir Vovk of Royal Holloway College, James Gleeson of University of Limerick, David Malone and Ken Duffy of the University of Maynooth for their comments on the Manuscript, and lastly Geoffrey Hinton, University of Toronto for bringing our attention to preceding work done in the Alphanets paper by John Bell \cite{B91}.

\ifCLASSOPTIONcaptionsoff
  \newpage
\fi



%

%
\begin{IEEEbiographynophoto}{Dr. Finn Macleod}
Phd University College Cork with Professor Alexei Pokrovski, Research Fellow University of Limerick. Dr Finn Macleod is Founder and Director of Beautiful Data (www.beautifuldata.eu)
\end{IEEEbiographynophoto}



\vfill


\end{document}